# A Method for Selecting Noun Sense using Co-occurrence Relation in English-Korean Translation

Hyonil Kim[a] and Changil Choe[b]

a.College of Computer Science, Kim Il Sung University, D.P.R.K

Email: hyonilkim@yahoo.com

b.Department of mathematics, Kim Il Sung University, D.P.R.K

Email: mathcci@yahoo.com

**Abstract**

The sense analysis is still critical problem in machine translation system, especially such as English-Korean translation which the syntactical different between source and target languages is very great. We suggest a method for selecting the noun sense using contextual feature in English-Korean Translation.

**1. Introduction**

In the machine translation, the WSD is generally one of the most difficult problems and many methods are already proposed. The method using knowledge base, such as collocation dictionary, and the example-based method are well-known. In the former method the sense of the words of the input sentence are estimated using the preparative knowledge about the mutually syntactically and semantically combinable words and therefore once the knowledge database is constructed the precision of the estimation is very high whereas the construction of the database is very expensive according to change of the domain. Also, this method didn't consider the indirect semantic relation between words in a sentence. In the example-based method the similarity between the input sentence and the sample in example base is estimated and the best similar sample is selected. This method has also serious problem that constructing the example base is very expensive and those usage efficiency is very low. To overcome these disadvantages the semantic hierarchy, such as thesaurus or WordNet is introduced, but it is yet trouble to machine translation.

## 2. Selecting the Noun Sense by Co-occurrence Relation

In the sentence the words have direct or indirect semantic relation with other words. In other words, the words which frequently occur in same sentence are very semantically related and the possibility of the co-occurrence of the words which don't have semantic relation is few. Therefore this relation plays important role in the WSD.

We regard the relation between those words which frequently occur in same sentence as co-occurrence relation. We represent the set of words which have co-occurrence relation with the word w as CWS(co-occurrence word set), and C(w) denotes the co-occurrence word set of w.

CWS of each word is extracted by statistically testing in English raw corpus. In this time the preposition, article, determiner, conjunction aren't considered.

Assume that M(w) is sense set of the word w, and then the problem which estimate optimal sense of the word, denoted as $m^*$, in the sentence s as follows:

$$m^* = \arg\max_{m \in \mathbf{M(w)}} P(w, m | \mathbf{s}) \qquad (1)$$

In the sentence, the words influence to estimate the sense of the word w, and so we can rewrite Eq. 1 as Eq. 2.

$$m^* = \arg\max_{m \in \text{mean}(w)} P(w, m | \mathbf{C(w)}) =$$
$$= \operatorname{argmax}_{m \in \mathbf{M(w)}} P(w, m) \cdot P(\mathbf{C(w)} | w, m) =$$
$$= \operatorname{argmax}_{m \in \mathbf{M(w)}} P(m|w) \cdot P(w) \cdot P(\mathbf{C(w)} | w, m) =$$
$$= \operatorname{argmax}_{m \in \mathbf{M(w)}} P(m|w) \cdot P(\mathbf{C(w)} | w, m) \qquad (2)$$

Assuming $P(C(w)|w, m) = \prod_{c \in C(w)} P(c|w, m)$, we can rewrite Eq. 2 as follows

$$m^* = \arg\max_{m \in \mathbf{M(w)}} P(m|w) \cdot \prod_{c \in \mathbf{C(w)}} P(c|w, m) \qquad (3)$$

As we are faced with data sparse problem in the natural language processing, it isn't also guaranteed to estimate correctly the parameter $P(c|w, m)$ in Eq. 3.

To solve this problem, we consider the hypernym which has a is-a relation with the sense assigned the Korean word in E-K dictionary and is found in the thesaurus or WordNet.

In this paper, we define ESS(Extended Sense Set) as the extension of the sense set, denote as $M(w)$, of the word w in E-K dictionary.

$$E(w) = M(w) \cup \{m' | isa(m, m') = true, m \in M(w),$$
$$\forall m'' \in M(w)(m \neq m''), isa(m'', m') = false \}$$

where $isa(m, m') = true$ means that the sense $m'$ is hypernym of the sense m.

Using this, we can rewrite Eq. 3 as follows:
$$\hat{m} = \arg\max_{m \in E(w)} P(m|w) \cdot \prod_{c \in C(w)} P(c|w, m) \qquad (4)$$

Because of the definition of the extended sense set, the solution $\hat{m}$ of Eq. 4 is an element of $E(w)$ and either an element or a hypernym of an element of $M(w)$. Therefore once the solution $\hat{m}$ of Eq. 4 is solved, we can estimate the sense $m^*$ of the word w.

### 3. Parameter Estimation

To estimate two parameters $P(m|w)$ and $P(c|w, m)$, we need sense tagged English corpus. But To build such a large training data is very expensive.

In this paper, we automatically align the sentences in E-K bilingual corpus and then automatically align the words using E-K bilingual dictionary, and then we extract the training data. In this time, the precision of automatic alignment is importantly considered, but the recall not.

Fig. 1 shows the process of estimating parameters in a diagram.

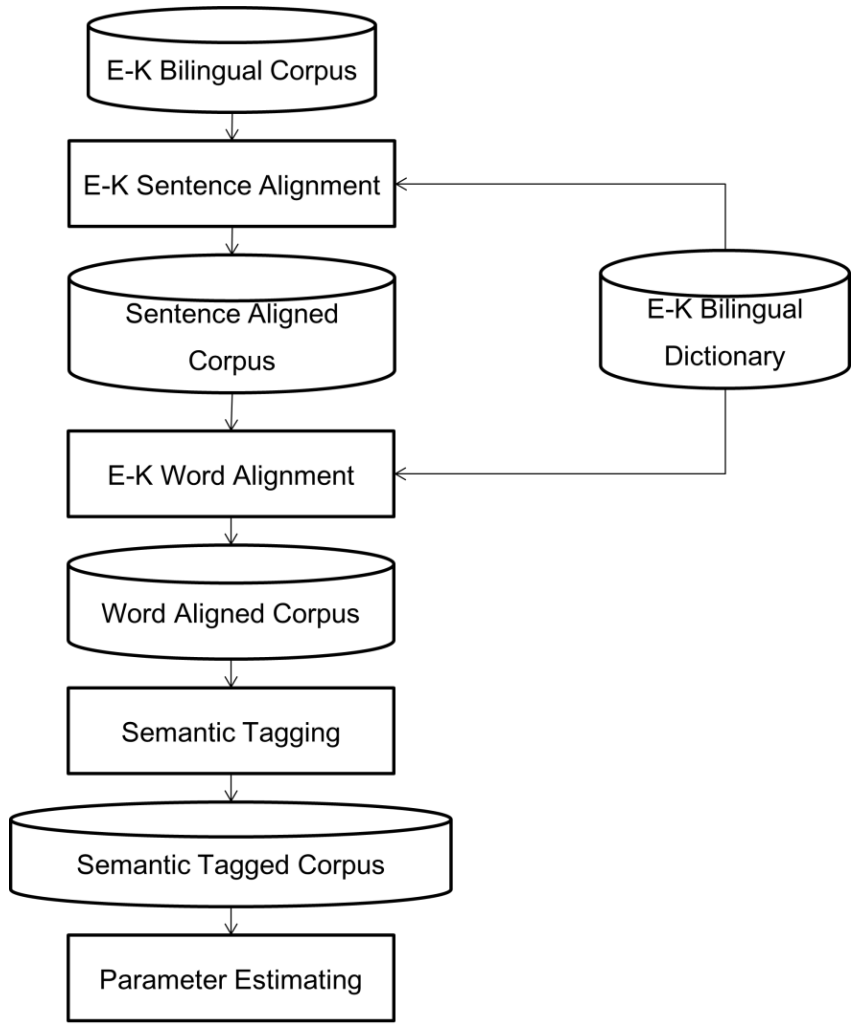

Fig. 1 Parameter estimating process

From the training data which is extracted in E-K bilingual corpus, we estimate two parameters as follows:

$$P(m|w) = \frac{\sum_{m'} CT(w,m') * WG(m,m') + 1}{CT(w) + N_1} \quad (5)$$

$$P(c|w,m) = \frac{\sum_{m'} CT(w,m',c) * WG(m,m') + 1}{\sum_{m'} CT(w,m') * WG(m,m') + N_2} \quad (6)$$

where CT(w) represents the count of the word w which is assigned semantic tag, and CT(w, m′) the count of the word w which is assigned semantic tag m′, and CT(w, m′, c) the count of the sentence in which the word assigned semantic tag m′ and the word c are appeared.

WG(m, m′) represents a weight which means the possibility of replacing the semantic tag m with virtual semantic tag m′. In this paper, we set this weight to be 1 in the case that m is equal to m′, to be a factor in the case that m′ is hypernym of m, and to be 0 in other case. $N_1$ and $N_2$ are the constants for the smoothing.

## 4. Experiments

In this paper we prepared the following data about two polysemy words "plant" and "bank" to test the effect of our approach.

Table 1 shows the sense set and ESS of two words "plant" and "bank" in the dictionary of E-K machine translation system RyongNamSan.

| English word | Sense set | ESS |
|---|---|---|
| plant | factory, plant, equipment | factory, work area, establishment, place, plant, living thing, equipment, goods, thing, lifeless thing |
| bank | enterprise, embankment, shore | Enterprise, organization, social collective, composition, abstract thing, embankment, flood control equipment, service establishment, public establishment, establishment, place, concrete thing, shore, far and near, space |

Table 1. The sense set and ESS of two words "plant" and "bank"

Table 2 shows respectively the CWS of two words "plant" and "bank".

| English word | Set of Co-occurrence words |
|---|---|
| plant | animal, soil, root, seed, transgenic, growth, gene, nutrient, crop, water, leaf, produce, tissue, power, heat production, food, cell, fruit,… |
| bank | loan, credit, financial, central, capital, risk, lending, deposit, fund, river, cod, inshore, offshore, boat, … |

Table 2. CWS of "plant" and "bank"

Table 3 shows the result of extracting the sentences which include the word "plant" or "bank" from semantic tagged English corpus. We used 80 per cent of total data to estimate parameters and 20 per cent to test our approach.

| English word | number of sentences | Number of sentences by sense | | |
|---|---|---|---|---|
| | | Sense | Num of sentences | Training data | Test data |
| plant | 24951 | Factory | 10098 | 8091 | 2007 |
| | | Plant | 13094 | 10475 | 2619 |
| | | Equipment | 1759 | 1395 | 364 |
| bank | 9393 | Enterprise | 5230 | 4180 | 1050 |
| | | Embankment | 1294 | 1031 | 263 |
| | | Shore | 2869 | 2303 | 566 |

Table 3. Analysis of sentences which include "plant" or "bank"

The result of estimating the sense of two words "plant" and "bank" from test data is shown in Table 4.

We measured the recall, precision, and F-Score by each sense. For a sense m, the recall and precision is calculated as follows:

$$\text{recall} = \frac{\text{the number of correct estimations to be m}}{\text{the number of total words which tagged with m}} \times 100$$

$$\text{precision} = \frac{\text{the number of correct estimations to be m}}{\text{the number of estimations to be m}} \times 100$$

In this paper, we also consider the weights of the recall and precision to be same, and so calculated the F-Score as follows:

$$\text{F} - \text{Score} = \frac{2 \cdot \text{recall} \cdot \text{precision}}{\text{recall} + \text{precision}}$$

To measure the effect of the extended sense set, we make two experiments. First experiment is to measure the performance of the system without extended sense set, and second experiment is to measure one with extended sense set.

The result of the first experiment is shown in Table 4. And second one in Table 5.

| English word | sense | Number of sentences | Num. of correct/num. of total estimations | Recall (%) | Precision (%) | F-Score |
|---|---|---|---|---|---|---|
| plant | Factory | 2007 | 987/1004 | 49.2 | 98.3 | 65.6 |
|  | Plant | 2619 | 1741/1763 | 66.5 | 98.8 | 79.5 |
|  | Equipment | 364 | 52/54 | 14.3 | 96.3 | 24.9 |
| bank | Enterprise | 1050 | 825/838 | 78.6 | 98.4 | 87.4 |
|  | Embankment | 263 | 12/12 | 4.6 | 100 | 8.73 |
|  | shore | 566 | 293/301 | 51.8 | 97.3 | 67.6 |

Table 4. The result of the first experiment

In tables 4-5, we present the number of words which are correctly estimated to be corresponding sense and words which are estimated to be one. Based on these result we calculated the recall, precision, and F-Score.

In the first experiment average recall is 56.9%, average precision 98.4% and F-Score 72.1, and in the second average recall is 96.1%, average precision 96.1% and F-Score 96.1.

| English word | sense | Number of sentences | Num. of correct/num. of total estimations | Recall (%) | Precision (%) | F-Score |
|---|---|---|---|---|---|---|
| plant | Factory | 2007 | 1908/1995 | 95.1 | 95.6 | 95.4 |
| plant | Plant | 2619 | 2541/2664 | 97.0 | 95.4 | 96.2 |
| plant | Equipment | 364 | 329/331 | 90.4 | 99.4 | 94.7 |
| bank | Enterprise | 1050 | 1031/1067 | 98.2 | 96.6 | 97.4 |
| bank | Embankment | 263 | 249/250 | 94.7 | 99.6 | 97.1 |
| bank | shore | 566 | 542/562 | 95.8 | 96.4 | 96.1 |

Table 5. The result of the second experiment

These experiments show that introducing the extended sense set results in the significant enhancement of average recall from 56.9% to 96.1%, but average precision fall down from 98.4% to 96.1%. However F-Score, overall measure, is greater from 72.1 to 96.1.

**5. Conclusion**

In machine translation, semantic analysis is very important component, but it is still very hard. The main reason of it is cost of building the database.

In this paper, we automatically build the training data from E-K bilingual corpus using E-K sentence alignment and word alignment. And using thesaurus or WordNet we introduced the extended sense set and so we enhanced the efficiency of training.

Using ESS, we achieved significant enhancement of recall as 39.2%. The precision of our approach is measured as 96.08%. It means that our approach is effective on WSD.